\documentclass[letterpaper, 10 pt, conference]{ieeeconf}  

\IEEEoverridecommandlockouts                              

\usepackage{cite}
\usepackage{amsmath,amssymb,amsfonts,amsthm,dsfont} 
\usepackage{tabularx,booktabs}
\usepackage{graphicx}
\usepackage{xcolor}
\newcommand\tab[1]{Table~\ref{#1}}
\newcommand\fig[1]{Fig.~\ref{#1}}
\usepackage[font={small}]{caption}
\usepackage{subcaption}
\usepackage[breaklinks=true, colorlinks, bookmarks=true, citecolor=black, urlcolor=violet,linkcolor=black]{hyperref}

\theoremstyle{definition}
\newtheorem{definition}{Definition}

\newcommand{\pa}{\partial}
\newcommand{\ba}{\begin{align}}
\newcommand{\ea}{\end{align}}

\newcommand{\R}{{\mathbb R}}
\newcommand{\E}{{\mathbb E}}

\newcommand{\N}{{\mathcal N}}

\DeclareMathOperator*{\diag}{diag}

\DeclareMathOperator*{\vech}{vech}

\newtheorem*{problem*}{Problem}

\newcommand{\calA}{{\cal A}}

\newcommand{\calF}{{\cal F}}

\newcommand{\calI}{{\cal I}}

\newcommand{\calN}{{\cal N}}

\newcommand{\calS}{{\cal S}}

\newcommand{\bfa}{\mathbf{a}}

\newcommand{\bff}{\mathbf{f}}

\newcommand{\bfm}{\mathbf{m}}

\newcommand{\bfo}{\mathbf{o}}
\newcommand{\bfp}{\mathbf{p}}
\newcommand{\bfq}{\mathbf{q}}

\newcommand{\bfs}{\mathbf{s}}

\newcommand{\bfu}{\mathbf{u}}
\newcommand{\bfv}{\mathbf{v}}

\newcommand{\bfx}{\mathbf{x}}
\newcommand{\bfy}{\mathbf{y}}
\newcommand{\bfz}{\mathbf{z}}

\newcommand{\bflambda}{\boldsymbol{\lambda}}
\newcommand{\bfmu}{\boldsymbol{\mu}}

\newcommand{\bfpi}{\boldsymbol{\pi}}

\newcommand{\bfxi}{\boldsymbol{\xi}}

\newcommand{\bbR}{\mathbb{R}}
\newcommand{\bbS}{\mathbb{S}}

\newcommand{\bbZ}{\mathbb{Z}}

\title{\LARGE \bf 
Learning Continuous Control Policies for Information-Theoretic \\Active Perception}

\author{Pengzhi Yang \and Yuhan Liu \and Shumon Koga \and Arash Asgharivaskasi \and Nikolay Atanasov
\thanks{We gratefully acknowledge support from NSF FRR CAREER 2045945.}%
\thanks{The authors are with the Department of Electrical and Computer Engineering, UC San Diego, 9500 Gilman Drive, La Jolla, CA, 92093, USA, {\tt\footnotesize \{peyang,yul139,skoga,aasghari,natanasov\}@ucsd.edu}.
An open-source implementation is available at \url{https://github.com/JaySparrow/RL-for-active-mapping}.}
}

\begin{document}

\maketitle
\thispagestyle{empty}
\pagestyle{empty}



\begin{abstract}
This paper proposes a method for learning continuous control policies for exploration and active landmark localization. We consider a mobile robot detecting landmarks within a limited sensing range, and tackle the problem of learning a control policy that maximizes the mutual information between the landmark states and the sensor observations. We employ a Kalman filter to convert the partially observable problem in the landmark states to a Markov decision process (MDP), a differentiable field of view to shape the reward function, and an attention-based neural network to represent the control policy. The approach is combined with active volumetric mapping to promote environment exploration in addition to landmark localization. The performance is demonstrated in several simulated landmark localization tasks in comparison with benchmark methods.
\end{abstract}


\section{Introduction}

Recent advances in embedded sensing and computation hardware and in simultaneous localization and mapping (SLAM) software have enabled efficient, reliable, real-time mapping of unknown and unstructured environments \cite{cadena2016past}. However, most robot mapping methods are passive in utilizing sensing information and do not consider optimizing the robot's motion to improve performance. Yet, planning the robot's sensing trajectory to improve the quality of acquired information \cite{placed2022survey} may play a critical role in challenging environments in applications such as search and rescue \cite{kumar2004robot}, security and surveillance \cite{grocholsky2006cooperative}, and wildfire detection \cite{julian2019distributed}. 

This paper proposes an approach to learn continuous control policies for active perception. We consider a robot equipped with an onboard sensor capable of detecting objects of interest (landmarks) within a limited field of view (FoV). The objective is to maximize the mutual information between the landmark states and potential future sensor observations given past sensory data and robot trajectory. With a sensor model that is linear in the landmark states and subject to Gaussian noise, the mutual information objective is related to the information matrix of a Kalman filter (KF) estimating the landmark states. To prevent a non-smooth reward function due to the limited FoV, we use a differentiable FoV formulation for reward shaping. Then, an exploration policy is learned using proximal policy optimization (PPO) \cite{schulman2017proximal} over a continuous control space with a network architecture using an attention mechanism to handle multiple landmarks. The proposed method is demonstrated in simulation in comparison to an open-loop optimization method and a policy with a different network architecture.

\begin{figure}
    \centering
    \includegraphics[width=\linewidth]{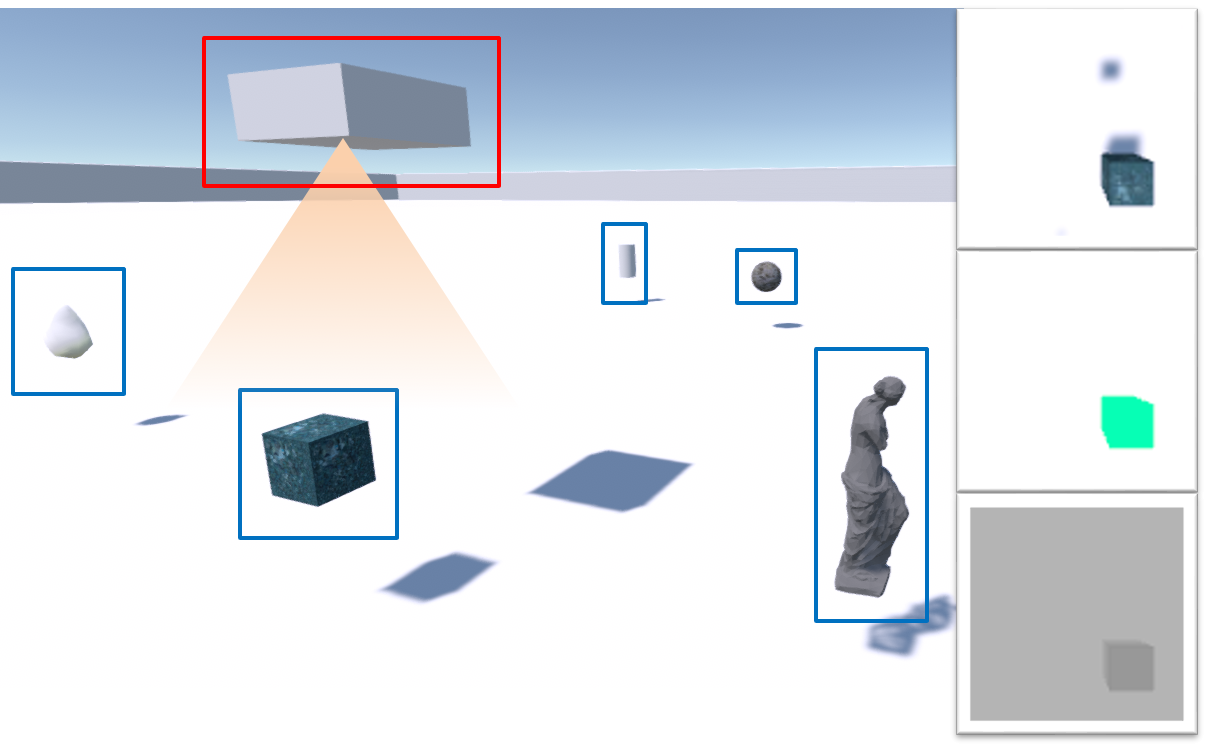}
    \caption{Active landmark localization in the Unity simulator \cite{juliani2018unity}. The left figure shows a third-person view of an agent (red box) with limited field of view (orange triangle) tasked with exploring and localizing landmarks (blue boxes) in the environment. The images on the right from top to bottom are RGB, semantic segmentation, and depth images obtained by the agent's onboard sensor.}
    \label{fig:Unity}
\end{figure}


Frontier-based exploration \cite{yamauchi1998frontier} is one of the early techniques for autonomous robot exploration. It drives the robot to map frontiers, separating explored and unexplored space. To accommodate sensing noise and uncertainty, information-based planning has been explored in many recent works including \cite{elfes1995robot,carlone2014active,atanasov2015decentralized,charrow2015information,zhang2020fsmi}. While many information-based exploration techniques assume a discrete control space composed of a finite number of action choices, \cite{koga2021active} proposed a gradient ascent approach, named iterative Covariance Regulation (iCR), which optimizes an information objective over continuous control space. The authors employed a differentiable FoV to obtain a differentiable objective. Because iCR provides an open-loop control sequence for a given environment, the solution cannot be applied to a new environment without online re-planning.


Learning a control policy from data is a central problem in reinforcement learning (RL) \cite{sutton2018reinforcement}. RL techniques coupled with deep learning representations have had impressive success in games \cite{mnih2015human}, where the action space is often discrete. More recently, deep RL algorithms have been developed for continuous control \cite{lillicrap2015continuous,schulman2017proximal}, which is necessary for various robotics tasks, including visual navigation for mobile \cite{tai2017virtual} and humanoid \cite{lobos2018visual} robots. In this paper, we consider learning a control policy for active perception as an alternative to view planning at execution time. Closely related to our work, Jeong et al. \cite{jeong2019learning} applied $Q$-learning to an active target tracking problem to maximize the mutual information between sensor data and the target states. Hsu et al. \cite{hsu2021scalable} developed a multi-agent version of \cite{jeong2019learning} by incorporating an attention-block in the Q-network architecture. Chen et al. \cite{chen2019learning} focused on learning exploration policies in a sample-efficient way by applying imitation learning first and then employing PPO for further improvement using a coverage reward. Julian and Kochenderfer \cite{julian2019distributed} used deep Q-learning for belief-space planning in a distributed wildfire surveillance scenario, modelled as a Partial Observable Markov Decision Process (POMDP). Chen et al. \cite{chen2020autonomous} tackled active exploration for landmark mapping using a graph neural network representing the exploration policy trained with deep Q learning and advantage actor critic methods.
Chaplot et al. \cite{chaplot2020learning} propose a modular and hierarchical approach to obtain a local policy by imitation learning from analytical path planners with a learned SLAM module and a global policy to maximize area coverage. Our work differs from \cite{chaplot2020learning} since we consider an information-theoretic objective with continuous control that does not rely on learning from any prior geometric planner. Lodel et al. \cite{lodel2022look} propose an information-theoretic objective to maximize conditional mutual information between a map estimate and new observations given past sensor data. The authors apply PPO to acquire reference view points that maximize the reward with local sensing of obstacles and the robot position. Our work differs from \cite{lodel2022look} in that we incorporate the posterior mean of the Kalman filter in the state and we use neural network attention to handle multiple landmarks.

The \textbf{contribution} of this paper is an approach for learning continuous control policies for active perception with information-theoretic reward, employing a differentiable FoV and an attention-based policy architecture. Our evaluation demonstrates that our method outperforms control policies with different neural network architectures and pre-computed exploration trajectories in a landmark localization problem and can also be utilized for simultaneous exploration of an unknown environment.

    
\section{Problem Statement}

This section formalizes active exploration and mapping as an optimal control problem. Consider a robot with state $\bfx_k \in \R^{n_x}$ and control input $\bfu_k \in \R^{n_u}$ at discrete time $k \in \bbZ_{\geq 0}$. 
The objective is to plan a trajectory to localize several landmarks $\bfy = [\bfy^{(1)}, \dots, \bfy^{(n_l)}]$, where $\bfy^{(j)} \in \R^2$ for $j \in \{1, \dots, n_l\}$ denotes the position of $j$-th landmark and $n_l$ is the total number of landmarks. We model the motion of the robot using deterministic nonlinear dynamics: 
%
\begin{align}
\bfx_{k+1} &= \bff(\bfx_k, \bfu_k) \label{eq:disc-model}, 
\end{align}
where $\bff: \bbR^{n_x} \times \bbR^{n_u} \to \bbR^{n_x}$ is a given function.  

Let $\calF \subset \R^2$ represent the FoV of the onboard sensors within the robot's body frame. The set of landmark indices within the FoV is:
\begin{align} \label{eq:IF} 
\calI_{{\calF}}(\bfx, \{\bfy^{(j)}\}) = \left\{ j \in \{1, \dots, n_l\} \mid   \bfq \left(\bfx, \bfy^{(j)} \right)  \in \calF \right\} 
\end{align}
where $\bfq(\bfx,\bfy^{(j)})$ returns the robot-body-frame coordinates of $\bfy^{(j)}$. 
Then, a sensor measurement is denoted by $\bfz_k = [ 
    \{ \bfz_k^{(j)}\}_{j \in \calI_{{\calF}}(\bfx_k, \{\bfy^{(j)}\})}
    ]\in \R^{n_z | \calI_{{\calF}}(\bfx_k, \{\bfy^{(j)}\})|}$ where $\bfz_k^{(j)} \in \bbR^{n_z}$ is an observation of $j$-th landmark with model:
\begin{align}
    \bfz_k^{(j)} &=  H(\bfx_k) \bfy^{(j)} +  \bfv_{k}, \quad \bfv_k \sim \N (0, V(\bfx_k)),  \label{eq:sensor}
\end{align}
for all $j \in  \calI_{{\calF}}(\bfx_k, \{\bfy^{(j)}\}),$ where the matrix $H(\bfx) \in \bbR^{n_z \times 2}$ captures the dependence of the observation on the robot state $\bfx$ and $V(\bfx) \in \bbR^{n_z \times n_z}$ is the sensing noise covariance.

We aim to maximize the conditional mutual information between the landmark states $\bfy$ and a new observation $\bfz_{k+1}$ conditioned on the past observations and robot states, denoted as $I(\bfy ; \bfz_{k+1} | \bfz_{1:k}, \bfx_{1:k+1})$. 
Due to the Gaussian sensor model \eqref{eq:sensor}, the conditional mutual information is given by:
\begin{equation}
    I(\bfy ; \bfz_{k+1} | \bfz_{1:k}, \bfx_{1:k+1}) = \frac{1}{2} \left( \log \det (Y_{k+1}) - \log \det ( Y_k)\right)
\end{equation}
where $Y_k \in \bbS_{\succ 0}^{2 n_{l} \times 2 n_l}$ is a symmetric positive-definite information matrix obtained by Kalman filter updates. Since the landmark measurements are independent, the information matrix is block-diagonal $Y_k = \diag(Y_k^{(1)}, \dots, Y_k^{(n_l)})$ with $Y_k^{(j)} \in \bbS_{\succ 0}^{2 \times 2}$. Due to the limited FoV sensor model, the update of the information matrix is applied only to the indices $j$ within the set $\calI_{{\calF}}(\bfx, \{\bfy^{(j)}\})$: 
\begin{align}
 Y_{k+1}^{(j)} & = Y_k^{(j)} + M(\bfx_{k+1}),  \label{eq:EIF-update} 
 \\
 M(\bfx) &:= H(\bfx)^\top V(\bfx)^{-1} H(\bfx).  \label{eq:sensor-info} 
\end{align}
Summarizing the formulation above, the active perception problem we address in this paper is presented below. 


\begin{problem*} 
Obtain a control policy $\pi$ that solves the infinite-horizon stochastic optimal control problem:
\begin{equation}\label{reward-def}
    \max_{\pi} \;\E_{\pi}\left( \sum_{k=0}^{\infty} \gamma^{k} \left( \log \det \left(  Y_{k+1} \right) - \log \det (Y_k) \right) \right),
\end{equation}
subject to \eqref{eq:disc-model}--\eqref{eq:sensor} and \eqref{eq:EIF-update}--\eqref{eq:sensor-info}, where the function $\pi$ maps a robot state $\bfx_k$ and measurement sequence $\bfz_{0:k}$ to a control input $\bfu_k$ and $\gamma \in [0,1)$ is a discount factor.
\end{problem*}





\section{Continuous Control for Active Perception}
\label{headings}



\begin{figure}
    \centering
    \includegraphics[width=0.95\columnwidth]{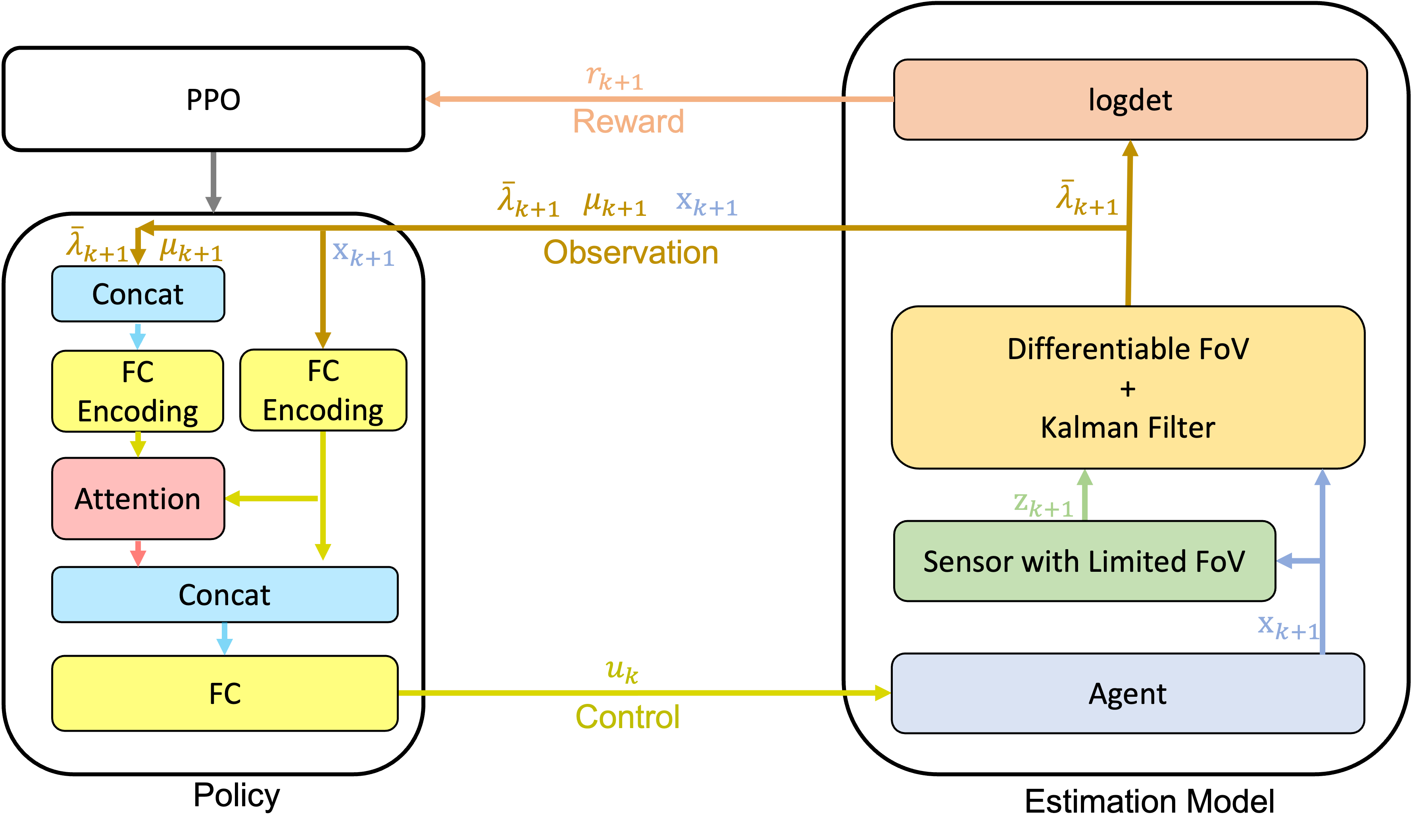}
    \caption{Estimation and control architecture used for active perception. A Kalman filter (right) updates the mean vector $\bfmu_k$ and vectorized information matrix $\bflambda_k$ of several landmarks of interest using a sensor observation $\bfz_{k+1}$ obtained from agent position $\bfx_{k+1}$. A reward function based on the $\log\det$ of the information matrix obtained with a differentiable field of view and the proximal policy optimization algorithm \cite{schulman2017proximal} are used to optimize the parameters of a control policy, whose representation (left) uses an attention mechanism to capture the relationship between the agent's position and the landmark states.}
    \label{fig:method-overview}
\end{figure}

This section presents our method for learning a continuous control policy to reduce the uncertainty in the landmark states. An overview of the method is shown in \fig{fig:method-overview}.

\subsection{MDP Formulation with Kalman Filter} 


The problem stated in the previous section is a POMDP \cite{kaelbling1998planning} since the landmark states are unobservable and the control policy should take the history of sensory data $\bfz_{1:k}$ into account. To avoid the POMDP complexity, we convert the problem into an equivalent MDP, by using a Kalman filter to compute a sufficient statistic of the state $\bfy_k$ given the observation history $\bfz_{0:k}$, namely the posterior distribution $\bfy_{k} | \bfz_{1:k} \sim \N\left(\bfmu_k, Y_k^{-1} \right)$. The posterior mean $\bfmu_k \in \bbR^{2 n_l}$ and covariance (inverse of the information matrix) $Y_k^{-1}$ are updated every time a new measurement is obtained. This setting is formalized as an MDP $(\calS, \calA, P, r, \gamma)$, with continuous state space $\calS$,  continuous action space $\calA$, transition probability distribution $P: \calS \times \calA \times \calS \to [0, 1]$, stage reward function $r: \calS \times \calA \to \R$, and discount factor $\gamma \in [0, 1)$. To define the MDP state variable, let $\bflambda_k := \vech(Y_k) \in \R^{n_{\lambda}}$ be the half-vectorization of the symmetric information matrix $Y_k$, which stacks the columns of the lower-triangular part of $Y_k$ into a vector with dimension $n_{\lambda} := n_l (2 n_l+1)$. Then, denote the Kalman filter update from time $k$ to $k+1$ for $j \in \calI_{{\calF}}(\bfx, \{\bfy^{(j)}\})$ as:
\begin{equation}\label{eq:KF-update}
[ \bfmu_{k+1}^{(j)}, \bflambda_{k+1}^{(j)} ] = KF(\bfmu_k^{(j)}, \bflambda_k^{(j)}, \bfz_{k+1},  \bfx_{k+1} ).
\end{equation}
We describe the design of the stage reward function $r$ next.

\subsection{Reward Shaping by Differentiable FoV} \label{differentiable_observation_model}


Typical RL methods for continuous control employ policy gradient methods \cite{lillicrap2015continuous}. However, due to the limited sensor FoV in \eqref{eq:IF}, the information update in \eqref{eq:EIF-update} may cause a sudden jump in the stage reward in \eqref{reward-def} with respect to the robot state and action. Such non-smooth behavior in the reward may prevent a policy gradient method from converging to a desired solution. To design a smooth reward function, we apply a differentiable FoV formulation \cite{koga2021active}, which serves as a reward shaping. Namely, the information update \eqref{eq:EIF-update} is performed for all $j \in \{1, \dots, n_l\}$, where the matrix $M(\bfx)$ in \eqref{eq:sensor-info} is replaced as follows: 
\begin{align}
    M(\bfx, \bfmu^{(j)}) &= \left(  1 - \Phi( d(\bfq(\bfx, \bfmu^{(j)}), {\mathcal F})) \right) M(\bfx),\label{eq:M} 
\end{align}
where 
$\Phi$ is a probit function \cite{bishop2006pattern}, defined by the Gaussian CDF $\Phi : \R \to [0, \;1]$,  $ \Phi(x) = \frac{1}{2} \left[ 1 + \textrm{erf} \left(\frac{x}{\sqrt{2} \kappa} - 2  \right) \right] $, where $\textrm{erf}(y):=\frac{2}{\sqrt{\pi}}\int_{0}^{y} e^{-t^2}dt$, and $d$ is a signed distance function associated with the FoV $\calF$ defined below.

%
 
\begin{definition}
The \emph{signed distance function} $d : \R^2 \to \R $ associated with a set $\calF \subset \bbR^2$ is:
\begin{align}
     d(\bfq, {\mathcal F})  = 
     \begin{cases}
     - \min_{\bfq^* \in \pa {\mathcal F}}  || \bfq - \bfq^*||, \quad \textrm{if} \quad \bfq \in {\mathcal F}, \\
     \phantom{-} \min_{\bfq^* \in \pa {\mathcal F}} ||\bfq - \bfq^*||, \quad \textrm{if} \quad \bfq \notin {\mathcal F},  
     \end{cases}
\end{align}
where $\pa {\mathcal F}$ is the boundary of ${\mathcal F}$.
\end{definition}

With the proposed differentiable FoV, the information vector $\bflambda_k$ in the Kalman Filter is replaced with the new variable $\bar \bflambda_k$ in the observation space. Overall, the MDP state vector $\bfs_k \in \R^{n_x + 2 n_l + n_{\lambda}}$ is defined as:
\begin{equation} \label{eq:land-state}
    \bfs_k = 
    \left[ 
    \begin{array}{ccc} 
    \bfx_{k}^\top & \bar \bflambda_k^\top 
    & \bfmu_k^\top
    \end{array}  
    \right]^\top
\end{equation}
and the action is $\bfa_k = \bfu_k$. The robot motion model in \eqref{eq:disc-model} and the KF update in \eqref{eq:KF-update} together with the differentiable FoV in \eqref{eq:M} define the MDP transition model. 

\subsection{PPO with Attention-Based Network Architecture}

The return from a state is defined as the sum of discounted future rewards, $R_k = \sum_{j=k}^{\infty} \gamma^{j-k} r(\bfs_j, \bfa_j)$. The value function $V^{\pi}: \calS \to \R$,  the action-value function $Q^{\pi}: \calS \times \calA \to \R$, and the advantage function $A^{\pi}: \calS \times \calA \to \R$ under a control policy $\pi$ are defined by $V^{\pi}(\bfs) =  \E_{\pi} \left[ R_0 | \bfs_0 = \bfs \right]$, $Q^{\pi}(\bfs, \bfa) = \E_{\pi} \left[ R_0 | \bfs_0 = \bfs, \bfa_0 = \bfa \right]$, $A^{\pi}(\bfs, \bfa) =  Q^{\pi}(\bfs, \bfa) - V^{\pi}(\bfs)$.
We apply PPO \cite{schulman2017proximal}, an actor-critic method for learning the neural network models of both the value function and the control policy. PPO has shown superior performance for continuous control tasks with smooth policy updates. The policy $\pi_{\theta}$ is updated to minimize a clipped surrogate objective: 
\small
\begin{align}
L_{\textrm{CLIP}} (\theta) = \E \left[ \min \left \{ \frac{\pi_{\theta}}{\pi_{\theta_{k}}} \hat A_k, \textrm{clip} \left(\frac{\pi_{\theta}}{\pi_{\theta_{k}}} , 1 - \epsilon, 1 + \epsilon \right) \hat A_k\right\} \right],  
\end{align}
\normalsize
where $\hat A_k$ is an estimate of the advantage function, set as $\hat A_k = \sum_{i = 0}^{T-k+1} \left( \gamma \lambda' \right)^i \delta_{k+i} $ where $\delta_k = r_k + \gamma V_{\phi}(s_{k+1}) - V_{\phi}(s_k)$, and $\lambda'$ represents the parameter controlling the balance between value estimations' bias and variance. The value function is subsequently updated to minimize $\sum_{k=0}^{T} (V_{\phi}(\bfs_k) - R_k)^2$. 

To take a reasonable trajectory that maximizes information gain, the agent should plan based on its previous observations. Some related works directly incorporate historical observations and agent states or use recurrent neural networks \cite{chen2020autonomous, lodel2022look}. In contrast, our model uses the landmark means and information matrices as part of the state, and because they form a sufficient statistic it is not necessary to have historical information.

Another important aspect of our design is that the agent should be aware of the relationship between its current position and the landmark states in order to prioritize observing uncertain landmarks. To capture the position relationship we employ an attention mechanism \cite{long2020evolutionary} in the design of the policy architecture. We encode the agent position $\bfx_k \in \mathbb{R}^{n_x}$ into $\textrm{Emb}(\bfx)$ with two 32-unit fully connected layers (FC), and the landmark states $\bfo^\textrm{lm}_k \in \mathbb{R}^{n_l \times 4}$ (reshaped from $[\bflambda_k^\top \ \bfmu_k^\top]^\top \in \mathbb{R}^{4 n_l}$) into  $\textrm{Emb}(\bfo^\textrm{lm}_k)$ with a 64-unit FC followed by a 32-unit FC. Then, we obtain the agent-landmark relationship embedding denoted as $\textrm{Emb}_{\textrm{RELP}}$ by calculating the attention of $\textrm{Emb}(x)$ over all $\textrm{Emb}(o^\textrm{lm}_k)$. Finally, we concatenate $\textrm{Emb}(x)$ and $\textrm{Emb}_{\textrm{RELP}}$ and pass through two 64-unit fully connected layers to compute the action or value. An illustration of our model is shown in \fig{fig:method-overview}.

\subsection{Joint Exploration and Landmark Localization} \label{sec:unified}

In addition to landmark localization, we consider simultaneous exploration of an unknown environment. A map state $ \bfm \in \R^{n_m}$ is defined by discretizing the space $\Omega \subset \bbR^2$ into $h_m \times w_m = n_m \in {\mathbb N}$ tiles, 
and associating each tile $j \in {1,\ldots,n}$ with a position $\bfp^{(j)} \in \R^2$ and an occupancy value $m^{(j)} \in \R$ ($m^{(j)} = 1$ if occupied, and $m^{(j)} = -1$ if free
). While Gaussian density estimate obtained from the KF provides a continuous real value, by applying a threshold function which returns $1$ for positive input and $-1$ for negative input, the map state can be estimated as a binary value. Using the index set notation of \eqref{eq:IF}, the set of map tile indices within the field of view can be described as $ \calI_{{\calF}}(\bfx, \{\bfp^{(j)}\})$. Using the image sensor from the flying robot, the occupancy within the FoV can be measured directly. The sensor state is given by $ \bfz_k^{\textrm{map}} = [\{  \bfz^{\textrm{map}}( m^{(j)})\}_{j \in \calI_{{\calF}}(\bfx_k, \{\bfp^{(j)}\})} ]$, and the sensor model is set as $\bfz^{\textrm{map}}( m^{(j)}) =   m^{(j)}+ v$, 
where $v \sim \calN(0, \sigma)$. Similarly to the landmark localization problem above, we consider a unified exploration and landmark localization problem by computing a control policy $\pi$, mapping a state $\bfx_k$ and observations $\bfz_{0:k} = [\bfz_{0:k}^{\textrm{land}\top}, \bfz_{0:k}^{\textrm{map}\top}]^\top$ to an input $\bfu_k$, that solves:
%
\begin{equation}\label{eq:Entropy} 
\max_{\pi}  \;\E_{\pi}\left(  \sum_{k=0}^{\infty} \gamma^{k}  \left( \rho \alpha^{\textrm{land}} r_k^{\textrm{land}}  + (1 - \rho ) \alpha^{\textrm{map}} r_k^{\textrm{map}} \right) \right),
\end{equation}
with $
r_k^{\textrm{map}} =  \log \det (\bar Y_{k+1}) - \log \det (\bar Y_k)$, 
$\bar Y_{k+1}^{(j)}  =\bar Y_{k}^{(j)} + \frac{1}{\sigma^2} \bfxi_k^{(j)}$, 
$\bfxi_k^{(j)} = 1 - \Phi( d(\bfq(\bfx_k, \bfp^{(j)}), {\mathcal F}))$, 
where $\bar Y_{k}^{(j)} \in \R_+$ is a $j$-th diagonal element in the information matrix $\bar Y_k$ of the volumetric map, subject to the models in \eqref{eq:disc-model}--\eqref{eq:sensor}. The hyper parameter $\rho \in [0, 1] $ controls the weight of each reward, and $\alpha^{\textrm{land}}, \alpha^{\textrm{map}}$ are normalization factors. One notable difference from the landmark scenario is that, the set of indices $\calI_{{\calF}}(\bfx, \{\bfp^{(j)}\})$ is independent from the map state $\bfm$, thereby the posterior mean of the Kalman Filter is not needed in the observation space.

The state vector $\bfs_k \in \R^{n_x+2n_l+n_{\lambda}+2n_m}$ consists of $\bfs_k^{\textrm{land}} \in \R^{n_x+2n_l+n_{\lambda}}$ of the landmark localization task \eqref{eq:land-state} and $\bfs_k^{\textrm{map}} \in \R^{2n_m}$ of the map exploration. The state $\bfs_k^{\textrm{map}}$ includes the agent's differentiable FoV $\bfxi_k^{(j)}$ and the current information of each pixel in the map $\bar Y_{k}^{(j)}$: $\bfs_k = \left[ \bfs_k^{\textrm{land}\top}; \bfs_k^{\textrm{map}\top} \right]^\top, \bfs_k^{\textrm{map}} = \left[ \{\bfxi_k^{(j)}\}_{j=1}^{n_m\top}  \{\bar Y_{k}^{(j)}\}_{j=1}^{n_m\top} \right]^\top$. The two states for landmark localization and volumetric exploration are processed independently to extract their corresponding feature vectors: $\bfs_k^{\textrm{land}}$ goes through the same attention-based network without the last fully-connected layer, while $\bfs_k^{\textrm{map}}$ is reshaped as a 2-channel image $S_k^{\textrm{map}} \in \R^{2 \times h_m \times w_m}$ and compressed by two convolutional layers followed by a flatten operation. The two feature vectors are concatenated and fed to two fully-connected layers to compute the action or value.

\section{Evaluation} 

We consider aerial surveillance of a 3-D environment using a flying robot with downward-facing RGB-D sensor. For simplicity, we assume that the orientation and height of the robot are not controlled, and the $x-y$ position of the robot can be controlled directly by linear velocities, while the landmarks are located in a 2-D plane with the same height. Hence, the robot state is the position $\bfx_k \in \R^2$ and the control input is the linear velocity $\bfu \in \R^2$ so that $\bfx_{k+1} = \bfx_k + \bfu_k$. The robot's FoV is a circle with a fixed radius on the ground. For sensing of landmarks, we suppose both range and bearing sensors are available, thereby the sensor model is described by the robot-body-frame coordinates, i.e, $H = I_2 $ and $V = \sigma^2 I_2$ with a sensor noise magnitude $\sigma \in \bbR_{ \geq 0}$. Under this setting, it is easily shown that the information matrix $Y_k$ by Kalman Filter \eqref{eq:EIF-update} and differentiable FoV \eqref{eq:M} becomes a diagonal matrix, and thus $\bflambda_k = \diag(Y_k) \in \bbR^{2 n_l}$ is sufficient for the information vector in observation space. At each episode, the initial estimations of landmark positions $\bfmu_0$ are sampled from a Gaussian distribution, while its mean value corresponds to landmarks' true positions, and the variance is determined by the sensor noise.

We demonstrate the performance of our approach in several simulations. First, we show qualitative and quantitative results for landmark localization. With non-uniform initial information values among the landmarks, the agent prioritizes landmarks with low information. Then, promising results for the joint exploration and landmark localization method are presented. Finally, we apply our method in a high-fidelity Unity simulation showing its potential for transfer in a real-world setting. 


\begin{figure*}
  \centering
  \resizebox{0.8\textwidth}{!}{
  \begin{tabular}{cccc}
      & \textbf{3 Landmarks} & \textbf{5 Landmarks} & \textbf{8 Landmarks}\\
     \textbf{Trajectories} & 
     \multicolumn{1}{m{5cm}}{\includegraphics[height=1.5in, trim={0.5cm .4cm .2cm .2cm}, clip]{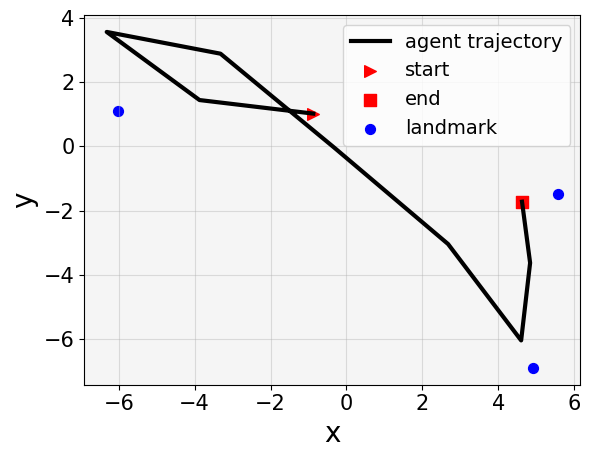}} & \multicolumn{1}{m{5cm}}{\includegraphics[height=1.5in, trim={1cm .4cm .2cm .2cm}, clip]{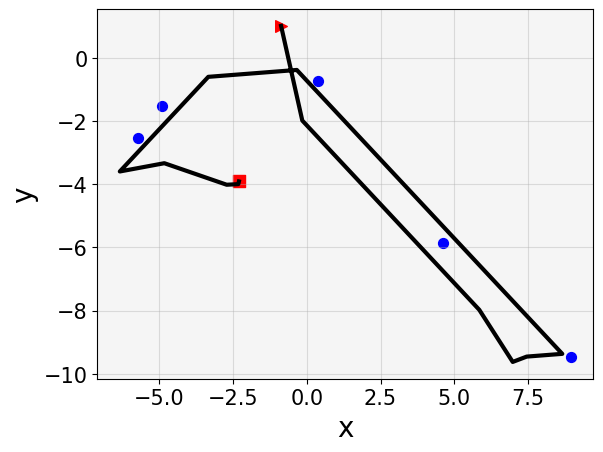}} & 
     \multicolumn{1}{m{5cm}}{\includegraphics[height=1.5in, trim={1cm .4cm .2cm .2cm}, clip]{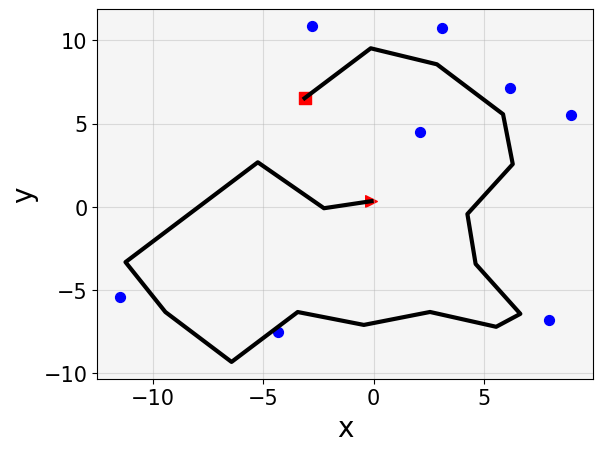}} \\
  \end{tabular}
  }
  \caption{Landmark localization trajectories. At test time, the agent is randomly initialized near the origin and has to explore the map to reduce uncertainty in the landmark positions with a limited sensing radius of $2$. The blue dots indicate the real positions of the landmarks. The small red triangle and square show the agent's initial and current positions, respectively.}
  \label{trajectories}
\end{figure*}

\subsection{Evaluation Settings} \label{sec:exp-setting}
We first conduct landmark localization experiments with $3$, $5$, and $8$ randomly scattered landmarks in the environment. Respectively, the agents takes $8$, $15$, and $18$ time steps per episode in each scenario during the training, so that the agent requires an efficient exploration by the terminal time. At the beginning of each episode, the agent's position is chosen randomly using a uniform distribution, $\bfx_0 \sim \textrm{Uniform}([ -2, 2] \times [-2, 2])$, while the landmark positions are specified using a uniform distribution within $[-8, 8] \times [-8, 8] $, $[-10, 10] \times [-10, 10] $ and $[-12, 12] \times [-12, 12]$, respectively. The control scaling factors are set as $3$, namely, the neural network stochastic control policy $\bfu_k \sim  \bfpi_{\theta}(\cdot | \bfs_k)$ is updated within the range $\bfu_k \in [- 3, 3] \times [-3, 3]$. 
Regarding the hyper-parameters in Sec.~\ref{differentiable_observation_model}, we chose a smoothing factor $\kappa$, sensing radius, and sensor noise $\sigma$ as $0.5$, $2$ and $0.5$,
respectively. The models are trained for a million time steps using a 24G NVIDIA GeForce RTX 3090 GPU. Specifically, the training process for each model was completed within 30 minutes in our experiment. 

\subsection{Landmark Localization Quantitative Comparisons} 
One demonstration of the trajectories generated from our learning policy in three different maps is depicted in \fig{trajectories}, which illustrates that the agent traverse through all the landmarks. We compare our method, named \textit{PPO-att}, to two baselines: (1) \textit{PPO-mlp}: a policy network replacing the attention block with a multi-layer perceptron and (2) \textit{iCR-landmark}: the open-loop optimal control method iCR \cite{koga2021active}, which performs finite-horizon trajectory optimization. We use the same hyper-parameters, protocols, and landmark and agent initial positions for all methods.

Each method is tested for 30 episodes per environment. Note that \textit{PPO-att} and \textit{PPO-mlp} are both trained with three random seeds to get three models in each environment. Each model runs $10$ episodes in each map. We utilize two metrics for quantitative comparisons: (1) \textit{Reward}: mean and standard deviation values of the cumulative reward, and (2) \textit{MAE}: the mean absolute error of the landmark localization at the end of each episode. Both metrics are computed over 30 episodes. We also tested the methods in environments with and without additive Gaussian noise in the robot dynamics \eqref{eq:disc-model}. The results are shown in \tab{Quantitative_Results}.

\textit{iCR-landmark} is capable of finding an optimal control sequence after sufficient number of iterations. However, it is not able to generalize the control input to different ground truth of the landmark positions, and its performance is highly dependent on a precomputed initial control sequence. The iCR-landmark trajectory can be recomputed at run-time, but such approach is out of our scope aiming to generalize a pre-computed policy to new environments. Secondly, according to the acquired data, \textit{PPO-mlp} performs relatively well when there are only three landmarks but obtains the worst results when there are more landmarks. Intuitively, although \textit{PPO-mlp} is able to handle landmark randomization, it cannot extract relationship features between landmarks and the agent. Thus, when the amount of the landmarks increases, exploration fails and leads to a significant performance drop. In contrast, our method always performs the best among other compared methods and is able to localize randomized landmarks with a small mean absolute error at the end of each episode. It is noteworthy that \textit{PPO-mlp} also exhibits an advantage over other methods in scenarios with agent motion noise. They demonstrated the robustness of the proposed method in noisy dynamical environments.

\begin{table}[h]
\Huge
\centering
\caption{{Quantitative Comparison Results.} The table shows the average and standard deviation for the cumulative episode rewards (higher is better) and the average estimation error after mapping (lower is better). Tests are performed in three kinds of maps with or without motion noise.}
\resizebox{1\columnwidth}{!}{%
    \begin{tabular}{cccccccc}
        \toprule
          Method & & \multicolumn{2}{c}{$\textrm{3 Landmarks}$} & \multicolumn{2}{c}{$\textrm{5 Landmarks}$} & \multicolumn{2}{c}{$\textrm{8 Landmarks}$} \\\cline{3-4}\cline{5-6}\cline{7-8}
           & & Reward & MAE & Reward & MAE & Reward & MAE \\\midrule
         \textit{iCR-landmark} & w/o noise & 11.93 $\pm$ 6.71 & 0.37 & 20.57 $\pm$ 7.58 & 0.34 &
         29.18 $\pm$ 10.0 & 0.36 \\
         & w/ noise & 13.3 $\pm$ 5.32 & 0.37 & 18.43 $\pm$ 6.48 & 0.39 & 25.31 $\pm$ 7.69 & \textbf{0.34}
         
         \\ \hline
         \textit{PPO-mlp} & w/o noise & 16.39 $\pm$ 4.82 & 0.34 & 18.97 $\pm$ 6.23 & 0.36 &
         16.36 $\pm$ 10.39 & 0.37 \\
         & w/ noise & 16.27 $\pm$ 4.9 & \textbf{0.32} & 15.26 $\pm$ 6.39 & 0.36 & 16.78 $\pm$ 9.64 & 0.37
         
         \\ \hline
         \textit{PPO-att} & w/o noise & \textbf{18.54 $\pm$ 2.9} & \textbf{0.29} & \textbf{30.27 $\pm$ 3.0} & \textbf{0.31} &
         \textbf{38.25 $\pm$ 8.0} & \textbf{0.35} \\
         & w/ noise & \textbf{18.13 $\pm$ 3.3} & \textbf{0.32} & \textbf{26.53 $\pm$ 6.17} & \textbf{0.33} &
         \textbf{30.15 $\pm$ 11.03} & 0.36
         \\
        \bottomrule
    \end{tabular}}
\label{Quantitative_Results}
\end{table}


\begin{figure*}
\begin{minipage}{0.49\linewidth}
  \resizebox{\linewidth}{!}{
  \begin{tabular}{cc}
      \multicolumn{1}{m{5cm}}{\includegraphics[height=1.5in, trim={.5cm .8cm .2cm .2cm}, clip]{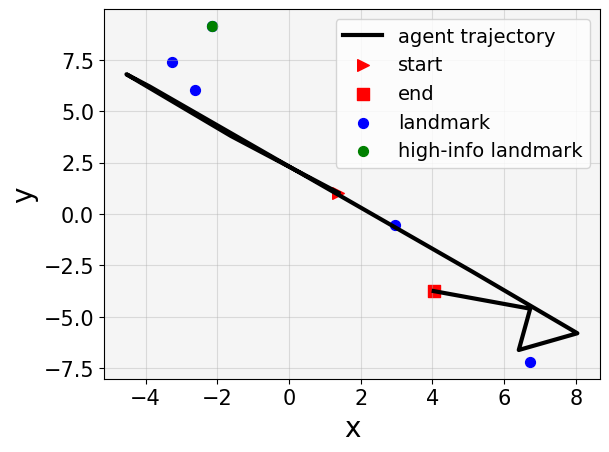}} & \multicolumn{1}{m{5cm}}{\includegraphics[height=1.5in, trim={1cm .8cm .2cm .2cm}, clip]{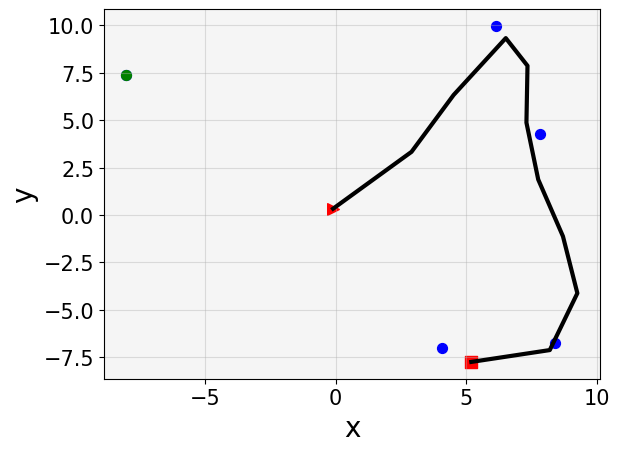}} \\
      \multicolumn{1}{m{5cm}}{\includegraphics[height=1.5in, trim={0cm .4cm .2cm .2cm}, clip]{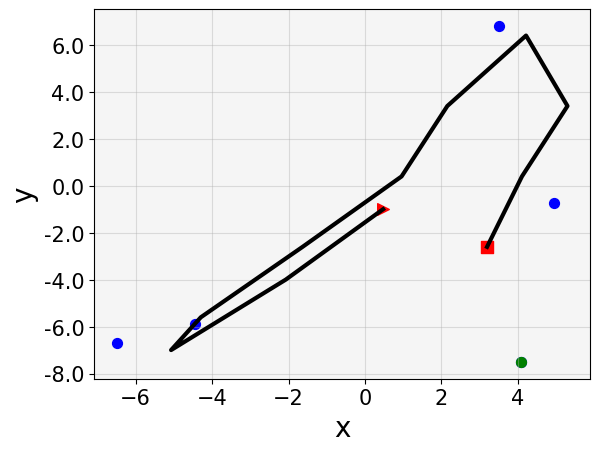}} & \multicolumn{1}{m{5cm}}{\includegraphics[height=1.5in, trim={1cm .4cm .2cm .2cm}, clip]{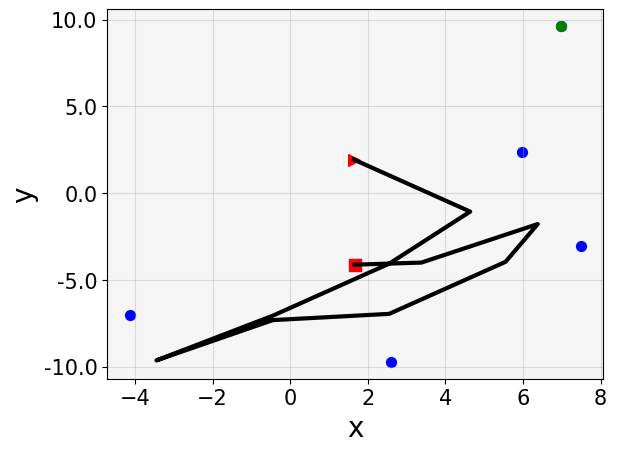}}\\
  \end{tabular}}
  \caption{Trajectories under non-uniform initial information values. When there is one landmark with higher initial information, the agent prioritizes the sensing of other landmarks with lower information. The green dots illustrate landmarks with higher information.}
  \label{prioritized_trajectory} 
\end{minipage}%
\hfill%
\begin{minipage}{0.5\linewidth}
    \includegraphics[width=\linewidth]{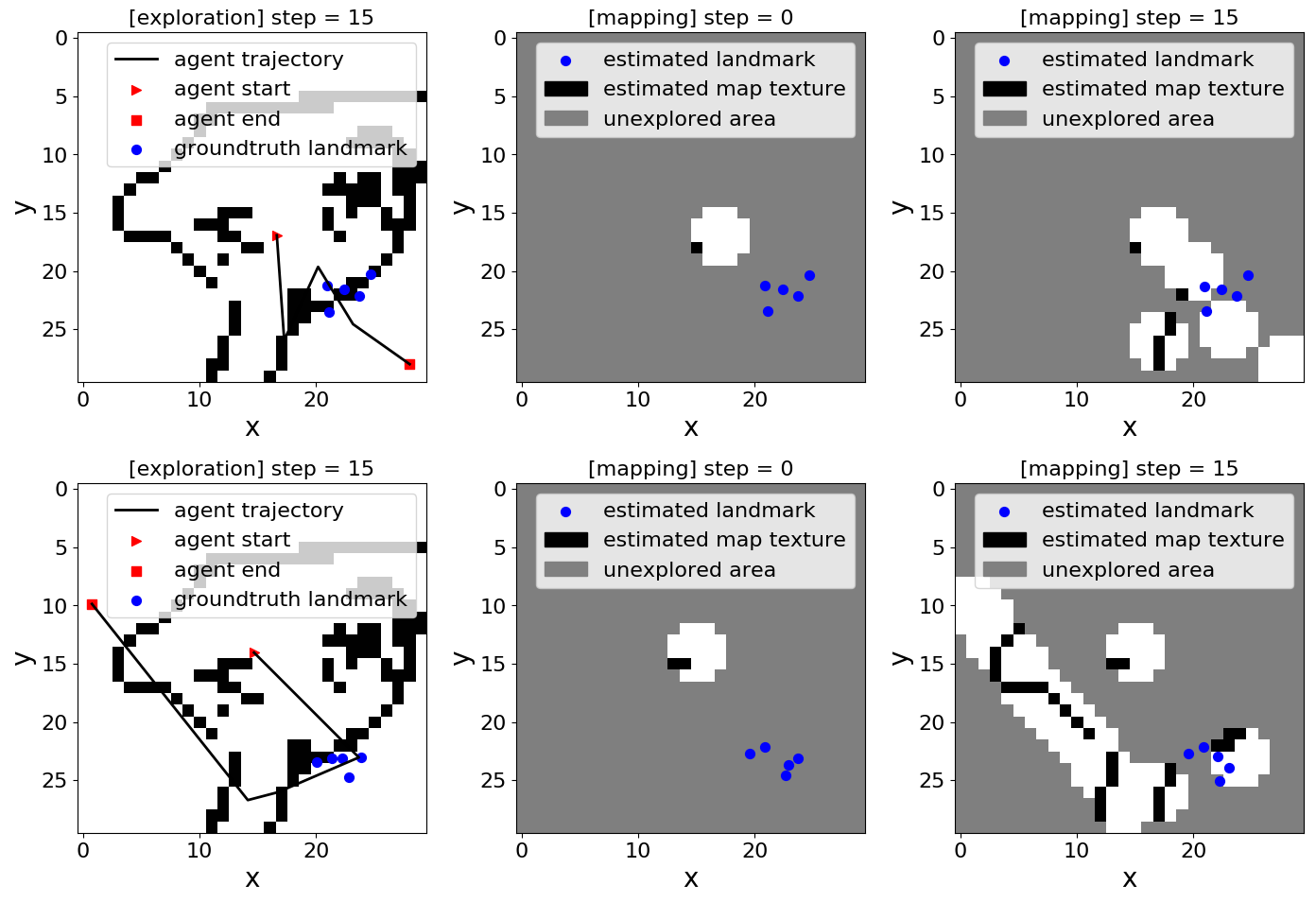}
    \caption{Landmark localization and exploration using an exploitation policy (top) and an exploration-exploitation policy (bottom). The left column shows the final trajectory on the ground-truth map. The two columns on the right show the estimated landmark positions and occupancy map at the beginning and the end of the episode.}
    \label{fig:volumetric-results}
\end{minipage}
\centering
\includegraphics[width=0.9\linewidth]{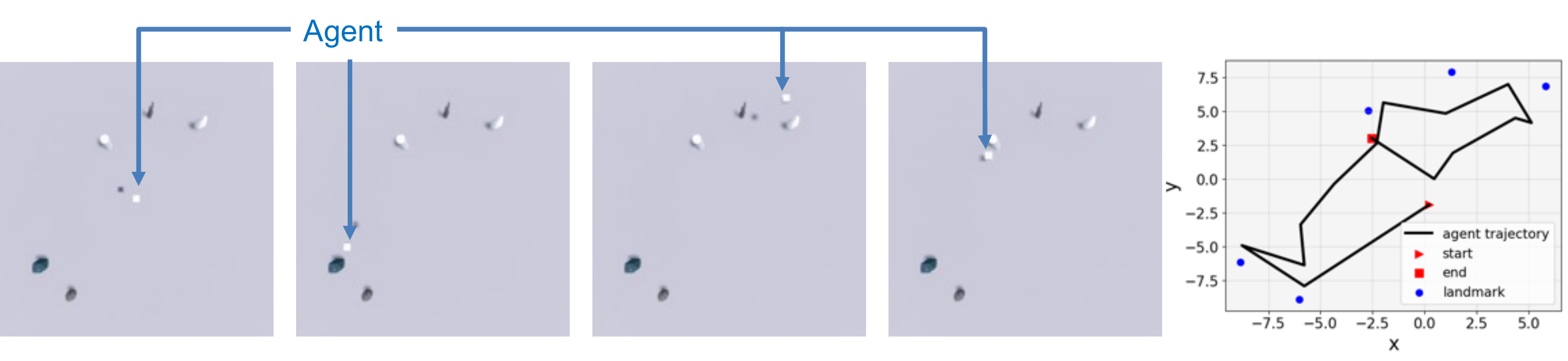}
\caption{Snapshots at time steps $0$, $5$, $10$ and $15$ in an active landmark localization episode in the Unity simulator.}
\label{fig:Unity-trajectory}
\end{figure*}

\subsection{Landmark Localization with Non-uniform Information}
We conduct a supplementary experiment to further demonstrate the features and advantages of our method by setting non-uniform initial information, where one landmark has much higher information (named \textit{high-info landmark} hereafter) than others. Fig.~\ref{prioritized_trajectory} shows different trajectories generated by the learned control policy for different landmark configurations. We can clearly see that the agent always prioritizes its exploration to landmarks with lower initial information values, and even ignores the \textit{high-info landmark} due to the limited number of time steps. With these experiments, we verify that our approach is capable of prioritizing less certain landmarks.

\subsection{Joint Exploration and Landmark Localization}


We also evaluate the joint exploration and landmark localization method described in Sec.~\ref{sec:unified}. The environment size is set as $30 \times 30$. During training, the agent is randomly initialized with a uniform distribution on $[13, 17]$ for both the $x$ and $y$ axes. The initial positions of the 5 landmarks are also uniformly randomized in a larger $x, y$ range of $[5, 25]$. The time horizon is fixed at 15 steps per episode. The rest of the environment and training parameters are the same in Sec.~\ref{sec:exp-setting}.

We compare two policies trained with different reward weights $\rho$ in \eqref{eq:Entropy}. The weight of the \emph{exploration-exploitation policy} is set to $\rho = 0.2$, i.e., the agent is pursuing both map exploration and landmark localization exploitation. The weight in the \emph{exploitation policy} is set to $\rho = 1.0$, i.e., the agent is only exploiting the landmark localization since only $r_k^{\textrm{land}}$ is kept in the reward.

To demonstrate the agent's ability of localizing the landmarks and exploring the map simultaneously, the 5 landmarks are randomly initialized in a more concentrated area (e.g. $[20, 25]$ for $x$ and $y$ in \fig{fig:volumetric-results}). Exploration-exploitation policy is expected to visit the landmarks first to obtain high information initially, and then explore the rest of the map to continue gaining information. In contrast, the exploitation policy is expected to remain around the landmarks leaving the majority of the map unexplored.

Fig.~\ref{fig:volumetric-results} shows the test results for the exploitation policy (top) and the exploration-exploitation Policy (bottom). It is obvious that the agent employing the exploration-exploitation policy continues to explore the map after detecting the landmarks, while with the exploitation policy the agent stays only around the landmarks. Qualitatively, the landmark positions are better estimated by the exploitation policy at the last step because it focuses on maximizing the information gain of only the landmarks.

\subsection{Unity Simulation} 
To examine the applicability of the method in real-world environments, we designed a Unity simulation \cite{juliani2018unity} which provides realistic training and testing settings. We used $5$ landmarks represented by different objects in Unity and positioned randomly in a bounded area in each episode. An aerial agent explores the environment to localize the scattered landmarks, as illustrated in \fig{fig:Unity}. The agent is equipped with a downward-facing aligned semantic and depth sensor with Gaussian noise on the depth values. When a landmark object is inside the sensor FoV, the agent uses the semantic map to extract the pixels of interest. Then, based on the depth values of the extracted pixels and the intrinsic and extrinsic camera matrices, we obtain a 3-D point cloud observation of object and estimate its center position in the plane. To guarantee the same shape of the sensor model, we also apply a squared signed distance function for the calculation. \fig{fig:Unity-trajectory} illustrates the agent exploring the environment. \textit{PPO-att} trained in a 2-D environment achieved a desired trajectory visiting all the landmarks.




\section{Conclusion} 

This paper proposed a learning method for active landmark localization and exploration with an information-theoretic cost over continuous control space. Key aspects of our method include (i) reward shaping using a differentiable FoV, (ii) attention-based neural network architecture for landmark prioritization, and (iii) joint landmark localization and environment exploration. Future work will focus on deterministic computation of reward gradients, collision and occlusion modeling, and deployment on a real robot system.



\bibliographystyle{IEEEtran}
\bibliography{BIB_ICRA23.bib}

\end{document}